\relax
\documentclass[letterpaper]{article} 
\usepackage{aaai22}  
\usepackage{times}  
\usepackage{helvet}  
\usepackage{courier}  
\usepackage[hyphens]{url}  
\usepackage{graphicx} 
\usepackage{natbib}  
\usepackage{caption} 
\DeclareCaptionStyle{ruled}{labelfont=normalfont,labelsep=colon,strut=off} 
\usepackage{stackengine}
\usepackage{multicol}
\usepackage{etoolbox}
\usepackage{times}
\usepackage{epsfig}
\usepackage{graphicx}
\usepackage{amsmath}
\usepackage{amssymb}
\usepackage{bm}
\usepackage{subfigure}
\usepackage{multirow}
\usepackage{color}
\usepackage{bm}
\usepackage{caption}
\usepackage[normalem]{ulem}
\useunder{\uline}{\ul}{}
\usepackage{verbatim}
\usepackage{microtype}
\usepackage{booktabs} 
\usepackage{makecell}
\usepackage{hhline}
\usepackage{array}
\usepackage{diagbox}
\usepackage{pifont}
\usepackage{algorithm,algcompatible,amsmath}
\usepackage{comment}

\usepackage{verbatim}
\usepackage{xspace}
\def\eg{\emph{e.g.}}

\def\ie{\emph{i.e.}}

\def\vs{\emph{vs.}}
\def\wrt{w.r.t.}

\usepackage{newfloat}
\usepackage{listings}
\floatstyle{ruled}
\newfloat{listing}{tb}{lst}{}
\floatname{listing}{Listing}
\title{Robust Physical-World Attacks on Face Recognition}
\author{
    Xin Zheng\textsuperscript{\rm 1}\equalcontrib,
    Yanbo Fan\textsuperscript{\rm 2}\equalcontrib,
    Baoyuan Wu\textsuperscript{\rm 3}\textsuperscript{\rm 4}\footnote{Corresponding author (wubaoyuan@cuhk.edu.cn).},
    Yong Zhang\textsuperscript{\rm 2},
    Jue Wang\textsuperscript{\rm 2},
    Shirui Pan\textsuperscript{\rm 1}
}
\affiliations{
    \textsuperscript{\rm 1} Monash University, Melbourne, Australia\\
    \textsuperscript{\rm 2} Tencent AI Lab, Shenzhen, China\\
    \textsuperscript{\rm 3} The Chinese University of Hong Kong, Shenzhen, China\\
    \textsuperscript{\rm 4} Shenzhen Research Institute of Big Data, Shenzhen, China
}

\usepackage{bibentry}

\begin{document}

\maketitle

\begin{abstract}
Face recognition has been greatly facilitated by the development of deep neural networks (DNNs) and has been widely applied to many safety-critical applications. However, recent studies have shown that DNNs are very vulnerable to adversarial examples, raising serious concerns on the security of real-world face recognition. In this work, we study sticker-based physical attacks on face recognition for better understanding its adversarial robustness. To this end, we first analyze in-depth the complicated physical-world conditions confronted by attacking face recognition, including the different variations of stickers, faces, and environmental conditions. Then, we propose a novel robust physical attack framework, dubbed PadvFace, to model these challenging variations specifically. Furthermore, considering the difference in attack complexity, we propose an efficient {\textit{Curriculum Adversarial Attack}} (CAA) algorithm that gradually adapts adversarial stickers to environmental variations from easy to complex. Finally, we construct a standardized testing protocol to facilitate the fair evaluation of physical attacks on face recognition, and extensive experiments on both dodging and impersonation attacks demonstrate the superior performance of the proposed method.
\end{abstract}
\section{Introduction}
Face recognition has achieved substantial success with the development of deep neural networks (DNNs) and has been widely applied to many safety-critical applications, such as video surveillance and face authentication \cite{huang2020curricularface,deng2019arcface,wang2018cosface}.
However, some recent works demonstrate that DNN-based face recognition models are very vulnerable to adversarial examples, even small and malicious perturbations can cause incorrect predictions \cite{sharif2016accessorize,sharif2019general,dong2019efficient,komkov2019advhat,xiao2021improving}.
For instance, when wearing an adversarial eyeglass frame, an attacker can deceive face recognition to be incorrectly recognized as another identity \cite{sharif2016accessorize}.
Such an adversarial phenomenon has raised serious concerns about the security of face recognition and it is imperative to understand its adversarial robustness.

Adversarial attack has been the most commonly adopted surrogate for the evaluation of adversarial robustness.
Existing attack methods on face recognition can be categorized into two types: (1) {\it digital attacks} where an attacker can perturb input images of face recognition directly in the digital domain \cite{goodfellow2014explaining,dong2019efficient,qiu2020semanticadv}, and (2) {\it physical attacks} realized by imposing adversarial perturbations to real faces in the physical world, \eg, wearable adversarial stickers \cite{sharif2016accessorize,sharif2019general,komkov2019advhat}.
As attackers usually cannot access and modify the digital input of physical-world face recognition systems, physical attacks are more practical for evaluating their adversarial robustness.
However, in contrast to the plethora of digital attack methods, few works are proposed to address physical attacks on face recognition, which remains challenging due to the complicated physical condition variations.

In this work, we study the sticker-based physical attacks that aim to generate wearable adversarial stickers to deceive state-of-the-art face recognition, for a better understanding of its adversarial robustness.
For robust physical attacks, an adversarial sticker should survive against complicated physical-world conditions.
To this end, we first provide an in-depth analysis of different physical-world conditions when conducting attacks on face recognition, including sticker and face variations, as well as environmental condition variations such as lighting conditions, camera angles, etc.
Then, we propose a novel robust physical attack framework, dubbed PadvFace, that considers and models these physical-world condition variations specifically.
Though some prior works show the possibility of performing physical attacks on face recognition \cite{sharif2016accessorize,sharif2019general,komkov2019advhat}, their performance is still unsatisfactory where only partial environmental variations are considered.
For instance, the work of \cite{komkov2019advhat} adopted {\it advhat} to attack physical-world face recognition systems. Yet it did not address the chromatic aberration of stickers, facial variations, as well as adequate environmental conditions. In this work, we demonstrate that these physical-world variations also influence attack performance significantly .

In terms of the optimization in physical attacks, {\it Expectation Over Transformation} (EOT) \cite{athalye2018synthesizing} is a standard optimizer that aggregates different physical-world condition variations to generate robust perturbations but simply treats each of them equally. However, we have the following observations: (1) the attack complexity of an adversarial sticker varies with different physical-world conditions,
and (2) the optimization of physical attacks generally lead to the non-convex optimization problem due to the high non-linearity of DNNs. Thus, simply adapting the adversarial sticker to all kinds of physical-world variations equally could make the optimization difficult and lead to inferior solutions.

To alleviate these, we propose a novel {\it Curriculum Adversarial Attack} (CAA) algorithm that advocates for exploring the attack complexity difference of physical-world conditions and gradually aggregates these conditions from easy to complex during the optimization. CAA adheres to the principles of {\it curriculum learning} \cite{bengio2009curriculum}, which has shown the benefits in obtaining better local minima and superior generalization for the non-convex optimization.
Finally, we build a standardized testing protocol for physical attacks on face recognition and conduct a comprehensive experimental study on the adversarial robustness of state-of-the-art face recognition models, under both dodging and impersonation attacks. Extensive experimental results demonstrate the superior performance of the proposed method.

The contributions of our work are four-fold:
\begin{itemize}
    \item We propose a novel physical attack method, dubbed PadvFace, that models complicated physical-world condition variations in attacking face recognition.
    
    \item We explore the attack complexity with various physical-world conditions and propose an efficient curriculum adversarial attack (CAA) algorithm.
    
    \item We build a standardized testing protocol for facilitating the fair evaluation of physical attacks on face recognition.
    
    \item We conduct a comprehensive experimental study and obtain the superior performance of physical attacks.
\end{itemize}

\section{Related Work}
Physical-world attacks aim to deceive deep neural networks by perturbing objects in the physical world \cite{athalye2018synthesizing,wang2019advpattern,xu2020adversarial}.
It is usually realized by firstly generating adversarial perturbations in the digital space and then fabricating and attacking in the physical world. 
Due to the complicated physical-world variations, there will be inevitable distortions when directly imposing the digital perturbations in the physical world.
Hence, the research focus of current physical attacks lies in how to efficiently model and incorporate complicated physical-world conditions.

Currently, most physical attacks focus on image classification \cite{duan2020adversarial,eykholt2018robust,zhao2020defenses,jan2019connecting,li2019adversarial} and object detection \cite{huang2020universal,zhang2018camou,chen2018shapeshifter,zhao2019seeing,zolfi2021translucent}.
However, face recognition is a quite different task with different model properties and objectives \cite{huang2020curricularface,deng2019arcface,wang2018cosface}. This makes attacking face recognition specifically different when conducting it in the physical world since more facial variations need to be involved beyond general environmental conditions. 
Recently, there have been some attempts of physical attacks on face recognition \cite{sharif2016accessorize,sharif2019general,komkov2019advhat,pautov2019adversarial,yin2021adv}.
Due to better reproducibility and being harmless to human beings, sticker-based attacks have been the mainstream approaches.

For sticker-based adversarial attacks, Sharif et al. \shortcite{sharif2016accessorize,sharif2019general} explored adversarial eyeglass frames for physical attacks.
They demonstrated that it was possible to deceive face recognition models by wearing an adversarial eyeglass. However, they did not involve the illumination or face variations. Furthermore, they only considered limited-scale face recognition models that were trained to recognize up to 143 identities and conducted attacks by perturbing face classification scores. In contrast, state-of-the-art (SOTA) face recognition models, such as ArcFace \cite{deng2019arcface} or CosFace \cite{wang2018cosface}, are based on the pair-wised (cosine) similarity and usually trained on tens of thousands of identities and millions of training images. Thus, such eyeglass-based attack methods fail to fool SOTA face recognition models, which is verified by Komkovand and Petiushko \shortcite{komkov2019advhat}.
The work of \cite{komkov2019advhat} proposed {\it advhat} to deceive the ArcFace model that trained on the large-scale dataset MS1MV2 \cite{deng2019arcface}. 
They verified the adversarial hat could reduce the cosine similarity of two facial images from the same person.
However, they did not consider the facial variations of attackers, as well as the chromatic aberration of adversarial stickers induced by printers and cameras.

Meanwhile, existing physical attacks commonly adopt the EOT optimizer \cite{athalye2018synthesizing} that treats different environmental variations equally during the optimization.
However, we demonstrate in this work that the attacking complexity varies with physical-world conditions and it is better to consider such characteristics for robust attacks.

\begin{figure*}[ht] 
	\centering
	\includegraphics[width=0.75\textwidth]{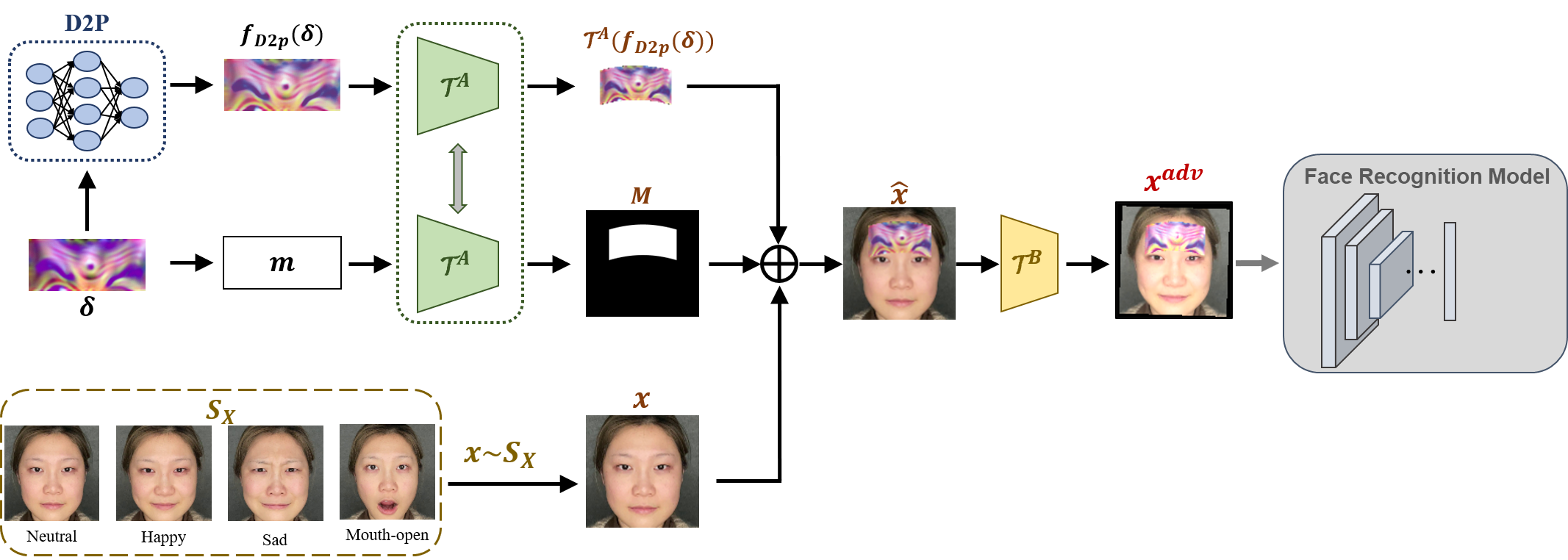} 
	\caption{Overall framework of the proposed robust PadvFace, where `D2P' denotes Digital-to-Physical Module, `{$\mathcal{T}^{A}$}' denotes Sticker Transformation Module, and `{$\mathcal{T}^{B}$}' conducts transformations on adversarial faces. } 
	\label{Fig.2}
\end{figure*}

\section{Proposed Method}

\subsection{Preliminary}
Let $\bm{x}$ be an input facial image and $\bm{x}^{a}$ be an anchor facial image, $f: \mathbb{R}^m \rightarrow \mathbb{R}^d$ be the face recognition model and $f(\bm{x}) \in \mathbb{R}^d$ denotes the learned feature embedding of $\bm{x}$.
The state-of-the-art face recognition model is generally realized based on the similarity between $f(\bm{x})$ and $f(\bm{x}^{a})$.
There are two types of adversarial attacks on face recognition: \textit{dodging attacks} that aim to reduce the similarity between facial images from the same identity and \textit{impersonation attacks} that aim to increase the similarity between facial images from different identities.
For dodging attacks where $\bm{x}$ and $\bm{x}^{a}$ are captured from the same identity, the optimization of adversarial sticker $\bm{\delta}$ can be formulated as 
\begin{equation} \textstyle
    \label{eq:digital attack}
    \min_{\bm{\delta}} \ \mathcal{L}_{sim}(f(\bm{x} + \bm{\delta}), f(\bm{x}^{a})),
\end{equation}
where $\mathcal{L}_{sim}$ denotes the attack loss, \eg, the cosine loss $\mathcal{L}_{sim} = {(1+cos(f(\bm{x} + \bm{\delta}), f(\bm{x}^{a})))}/{2}$. 
In contrast, impersonation attacks, where $\bm{x}$ and $\bm{x}^{a}$ are sampled from two different identities, can be optimized by minimizing $\mathcal{L}_{sim} = {(1-cos(f(\bm{x} + \bm{\delta}), f(\bm{x}^{a})))}/{2}$.

\subsection{Physical Attack Challenges}
The physical-world condition variations are challenging when attacking face recognition. Firstly, there would be {physical-world variations \wrt\ the sticker}: 1) spatial constraints that the sticker cannot cover all parts of faces, such as facial organs;
2) inevitable deformation and position disturbance when wearing the sticker and fitting it to the real face.
3) chromatic aberration of the sticker caused by the printers and cameras.
The sticker is first fabricated by printers and then wore and photographed by cameras for attacking face recognition.
Due to the limitation of printer resolution and different shooting conditions of photographing, there would be the chromatic aberration of the sticker between the digital space and the physical world.

Secondly, there would be {physical-world variations \wrt\ the adversarial face}: 1) photographing variations containing camera angles, head poses, and lighting conditions, etc.; 2) internal facial variations of attackers, such as different facial expressions and movements.

\subsection{Robust PadvFace Framework}
In this section, we propose a robust physical attack framework on face recognition, dubbed PadvFace, which considers and models the challenging physical-world conditions. 
Specifically, we adopt a rectangular sticker $\bm{\delta}$ pasted on the forehead of an attacker without covering facial organs. The overall framework of the proposed PadvFace is illustrated in Fig.~\ref{Fig.2}.

The rectangular sticker $\bm{\delta}$ is firstly fed into a Digital-to-Physical (D2P) module $f_{D2P}$ to model the chromatic aberration induced by printers and cameras. Then, a sticker transformation module $\mathcal{T}^A$ is introduced to simulate variations w.r.t. the sticker $\bm{\delta}$ when pasting on a real-world face. In the meantime, an initial mask $m$ is also fed into $\mathcal{T}^A$ by sharing the transformation with that on $\bm{\delta}$, generating the blending mask $M$. After these, the sticker is blended with a randomly selected facial image $\bm{x}$ according to the blending mask $M$, resulting in an initial adversarial image $\hat{\bm{x}}$. This initial adversarial image further inputs a transformation module $\mathcal{T}^{B}$ to simulate the environmental variations such as different poses and lighting conditions, leading to the ultimate adversarial facial image $\bm{x}^{adv}$ for deceiving face recognition, \ie, 
\begin{equation}
    \small
    \label{eq:transformation}
    \bm{x}^{adv} = \mathcal{T}^{B} ( \ {(1-M) \circ \bm{x} + M \circ \mathcal{T}^{A} (f_{D2P}(\bm{\delta}))}) + \bm{v},
\end{equation}
where $\bm{v}\sim \mathcal{N}(\mu,\sigma)$ is a random Gaussian noise and $M = \mathcal{T}^{A}(m)$, $\bm{x}$ is a randomly selected facial image.
More details of each module are introduced as follows.

\smallskip

\begin{figure}[b] 
	\centering
	\includegraphics[width=0.35\textwidth]{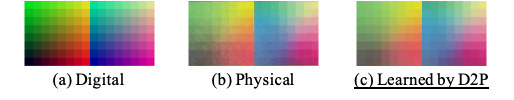} 
	\caption{Examples of color palettes.}
	\label{Fig.D2P}
\end{figure}

\textbf{Digital-to-Physical (D2P) Module.}
There are two types of chromatic aberration:
1) the fabrication error induced by printers, which mainly refers to the color deviation between the digital color and its printed version due to the limitation of printer resolution;
2) {photographing error} caused by cameras, such as sensor noises and different lighting conditions. 
To alleviate these issues, we develop a Digital-to-Physical (D2P) module to simulate the chromatic aberration of the sticker inspired by \cite{xu2020adversarial}. Specifically, the proposed D2P module is realized by training a multi-layer perception (MLP) to learn a 1:1 mapping from a customized digital color palette to its physically printed and photographed version, as illustrated in Fig.~\ref{Fig.D2P} (a-b), where Fig.~\ref{Fig.D2P} (c) presents the learned color palette from the proposed D2P module. Note that only parts of these color palettes are shown for the brief illustration, and the full color palettes and details of D2P are provided in \textit{Appendix}.

\textbf{Sticker Transformation Module $\mathcal{T}^A$} contains:
1) sticker deformation when pasting on a real-world face, including the off-plane bending and 3D rotations.
Following \cite{komkov2019advhat}, we adopt a parabolic transformation operator to simulate the off-plane bending and a 3D transformation for rotations;
2) position disturbance as it is hard to paste the sticker precisely at the same position as designed, such as random rotations and translations.

\textbf{Face Transformation Module.}
As analyzed above, the attacker also undergoes a set of physical-world variations induced by different poses, lighting conditions, and internal facial variations, etc.
To model these variations, we sample facial images from both physical and synthetic transformations.
Specifically, for internal facial variations, we capture real-world facial images with different facial expressions and movements, leading to a set of facial images $\mathcal{S}_{\bm{x}}$.
To simulate the variations induced by different camera angles, poses and lighting conditions, we consider transformation $\mathcal{T}^B$ that includes random rotation, scaling, translation, contrast and brightness on the adversarial facial images. 

\begin{table}[t]
    \centering
    \begin{small}
    \resizebox{0.45\textwidth}{!}{
    \begin{tabular}{cl}
    \toprule
    Module & Variations\\ \midrule
    D2P & Chromatic aberration from printers and cameras\\
    $\mathcal{T}^A$ & Parabolic transformation, rotation, translation\\
    $\mathcal{T}^B$ & Rotation, scaling, translation, contrast, brightness \\
    $\mathcal{S}_{\bm{x}}$ & Facial expressions, facial movements\\
    $\bm{v}$ & Random Gaussian noise\\
    \bottomrule
    \end{tabular}}
    \end{small}
    \caption{Physical-world variations in the proposed PadvFace.}
    \label{tab:Factors}
    \vskip -0.15in
\end{table}

The overall physical-world variations considered by the proposed PadvFace are summarized in Table~\ref{tab:Factors}. To address these challenging variations, EOT algorithm \cite{athalye2018synthesizing} is commonly used to implement robust physical attacks. Specifically, EOT first samples $n$ transformations $\mathcal{K} = \left\{(\bm{x}_i, \tau^{A}_i, \tau^{B}_i, \bm{v}_i)\right\}_{i=1}^{n}$ from $(\mathcal{S}_{\bm{x}}, \mathcal{T}^A, \mathcal{T}^B, \bm{v})$ and then optimizes the following objective as
\begin{equation}
    \small
    \label{eq:experical objective}
    \min_{\bm{\delta}} \frac{1}{n}\sum\nolimits_{k_i \in \mathcal{K}} \mathcal{L}_{sim,k_i}  + \alpha \mathcal{L}_{TV}(\bm{\delta}),
\end{equation}
where we adopt $k_i \triangleq (\bm{x}_i, \tau^{A}_i, \tau^{B}_i, \bm{v}_i)$ and $\mathcal{L}_{sim,k_i} \triangleq \mathcal{L}_{sim}(f(\bm{x}^{adv}_i | \bm{x}_i, \bm{\delta}, M_i, \tau^{A}_i, \tau^{B}_i, \bm{v}_i), f(\bm{x}^{a}))$ for shorthand.
$\bm{x}^{adv}$ is the corresponding adversarial image generated by Eq.~\eqref{eq:transformation}.
$\mathcal{L}_{TV}(\bm{\delta}) = \sum_{i,j} \left( (\bm{\delta}_{i,j} - \bm{\delta}_{i+1,j})^2 + (\bm{\delta}_{i,j} - \bm{\delta}_{i,j+1})^2 \right)^{\frac{1}{2}}$ is the total-variation loss introduced to enhance the smoothness of the sticker and $\alpha >0$ is a regularization parameter.
The D2P module and the synthetic transformations in $\mathcal{T}^A$ and $\mathcal{T}^B$ are all differentiable. And model \eqref{eq:experical objective} can be solved by stochastic gradient descent algorithm.

\begin{algorithm}[b]
    \caption{Curriculum Adversarial Attack}
    \label{alg: reg}
    \begin{small}
    \begin{algorithmic}[1]
    \REQUIRE  Attacked face recognition model $f$, physical transformations $\mathcal{K}=\left\{ {{k_i}} \right\}_{i = 1}^n$, anchor image $\bm{x}^{a}$, initial sticker $\bm{\delta}$, curriculum parameters $\left\{\lambda_t\right\}_{t=1}^T$ with $\lambda_1 < \lambda_2 < \cdots < \lambda_T$.
    \FOR{$t$ = $1,...,T$}
    \FOR{$e$ = $1,...,E$}.
    \STATE Fixed $\bm{\delta}$, updating $\bm{p}$ as $p_i = 1- \mathcal{L}_{sim, k_i}/\lambda_t$.
    \STATE Fixed $\bm{p}$, updating $\bm{\delta}$ via gradient descent.
    \ENDFOR
    \ENDFOR
    \ENSURE Robust adversarial sticker $\bm{\delta}$.
    \end{algorithmic}
    \end{small}
\end{algorithm}
\vskip -0.2in
\subsection{Curriculum Adversarial Attack}
\begin{figure}[t] 
	\centering
	\includegraphics[width=0.41\textwidth]{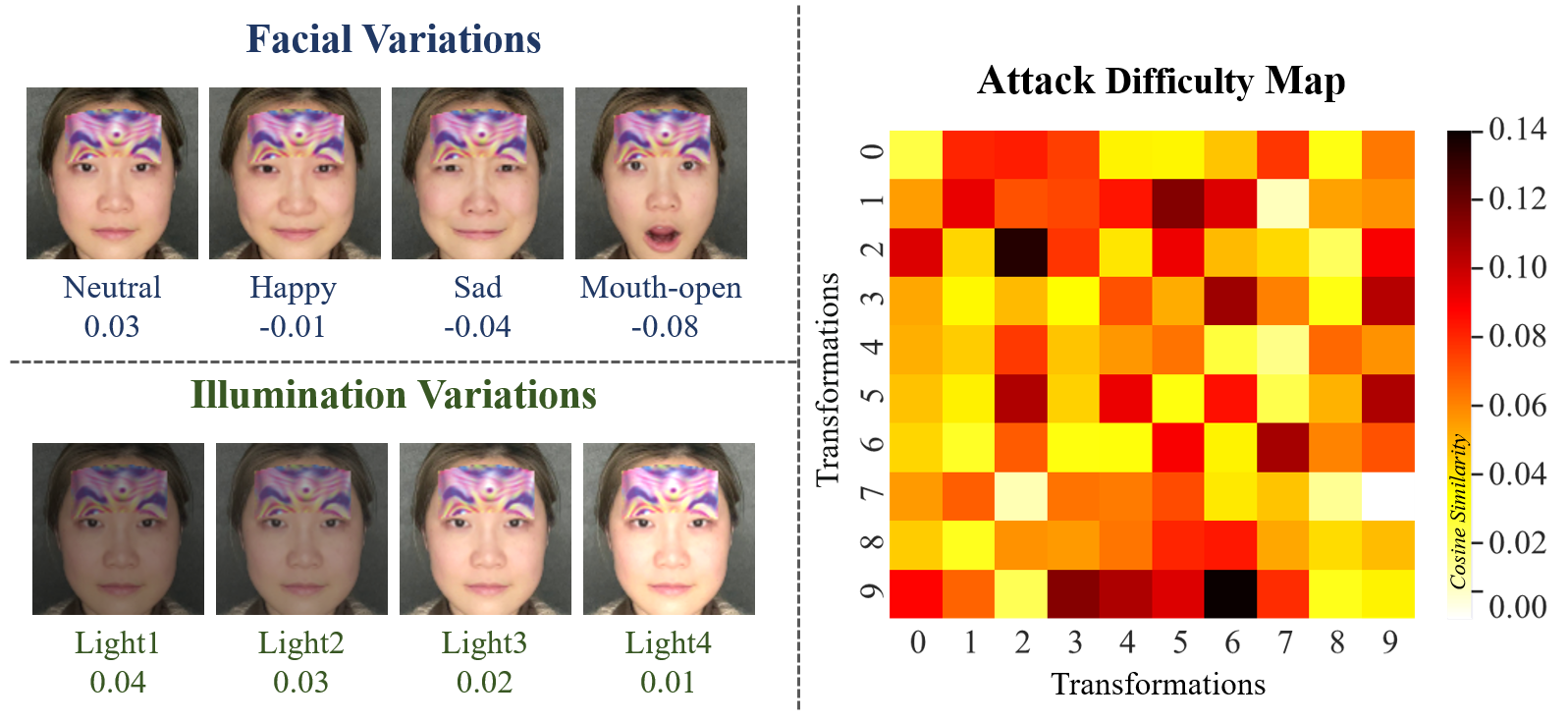}
	\caption{Dodging attack difficulties under different physical conditions, where metric is the adversarial cosine similarity.}
	\label{Fig.Trans_css}
	\vskip -0.1in
\end{figure}

We have the following observations for model \eqref{eq:experical objective}.
Firstly, for generating an adversarial sticker, the attack complexity varies with different physical-world conditions.
In Fig.~\ref{Fig.Trans_css}, we show the dodging attack performance of a fixed sticker under various facial variations, illumination variations, and 100 random sampled transformations from $(\mathcal{T}^A, \mathcal{T}^B)$.
And higher adversarial cosine similarity indicates lower attack performance and higher attack complexity.
The results show that the difficulty of physical attacks varies with different conditions.
Secondly, due to the high nonlinearity of DNNs, model \eqref{eq:experical objective} generally leads to a non-convex optimization problem.
Hence, directly fitting the adversarial sticker to all kinds of physical-world conditions $\mathcal{K}$ in model \eqref{eq:experical objective} could make the optimization difficult, resulting in inefficient solutions.

In light of these, we propose an efficient {\it curriculum adversarial attack} (CAA) algorithm to gradually optimize adversarial stickers from easy to complex physical-world conditions. Given an adversarial sticker $\bm{\delta}$, larger attack loss of $\mathcal{L}_{sim, k_i}$ indicates higher attack complexity under the condition $k_i$.
Thus, $\mathcal{L}_{sim, k_i}$ can serve as an appropriate surrogate for the measurement of the complexity of $k_i$.
Based on this, we assign a learnable weight parameter ${p}_{i} \in [0,1]$ for each transformation in $\mathcal{K}$ and formulate the objective of CAA as
\begin{equation} 
    \label{eq:spl objective}
    \min_{\bm{\delta}, p_i \in [0,1]} \frac{1}{n}\sum\limits_{k_i \in \mathcal{K}} \left\{ p_i \mathcal{L}_{sim,k_i} + \lambda g(p_i) \right\} + \alpha \mathcal{L}_{TV}(\bm{\delta}),
\end{equation}
where $g(p_i)=\frac{1}{2} p_i^2-p_i$ is a regularizer and $\lambda > 0$ is a curriculum parameter.

Let $\bm{p}=[p_1,p_2,\cdots,p_n]$, model \eqref{eq:spl objective} can be solved by alternatively optimizing $\bm{\delta}$ and $\bm{p}$ while keeping one of them fixed. 
Firstly, given $\bm{p}$, the optimization \wrt\ $\bm{\delta}$ reduced to
\begin{equation} \textstyle
    \label{eq:spl-v-step}
    \min_{\bm{\delta}} \frac{1}{n}\sum\nolimits_{k_i \in \mathcal{K}} p_i \mathcal{L}_{sim,k_i} + \alpha \mathcal{L}_{TV}(\bm{\delta}),
\end{equation}
which can be solved by stochastic gradient descent method. Secondly, given $\bm{\delta}$, the optimal $p_i^*$ is determined by
\begin{equation} \textstyle
    \label{eq:spl-p-step}
    \min_{p_i \in [0,1]} p_i \mathcal{L}_{sim,k_i} + \lambda (\frac{1}{2} p_i^2-p_i),
\end{equation}
leading to a closed-form solution as $p_i^ *  = 1 - \frac{\mathcal{L}_{sim,k_i}}{\lambda}$.
Thus, easier transformation with lower attack loss $\mathcal{L}_{sim}$ will be assigned with a larger weight $p$ and dominate the updating of $\bm{\delta}$ in the following step.
On the other side, the value of $\lambda$ is monotonically increased to involve more and more complex transformations during the optimization.
As a result, the proposed CAA algorithm can generate the sticker $\bm{\delta}$ from easy to complex transformations gradually. The overall algorithm of CAA is reported in Algorithm 1.

CAA adheres to the principle of curriculum learning \cite{bengio2009curriculum}, which learns from easy to complex tasks and has shown the benefits in obtaining better local minima and superior generalization for many non-convex problems \cite{huang2020curricularface,kumar2010self,fan2017self,cai2018curriculum}.
To the best of our knowledge, we are the first to explore the complexity of physical-world conditions in adversarial attacks and to aggregate them via curriculum learning towards the robust performance.

\section{Experiments}

\subsection{Testing Protocol}
\begin{figure}[t] 
	\centering
	\includegraphics[width=0.32\textwidth]{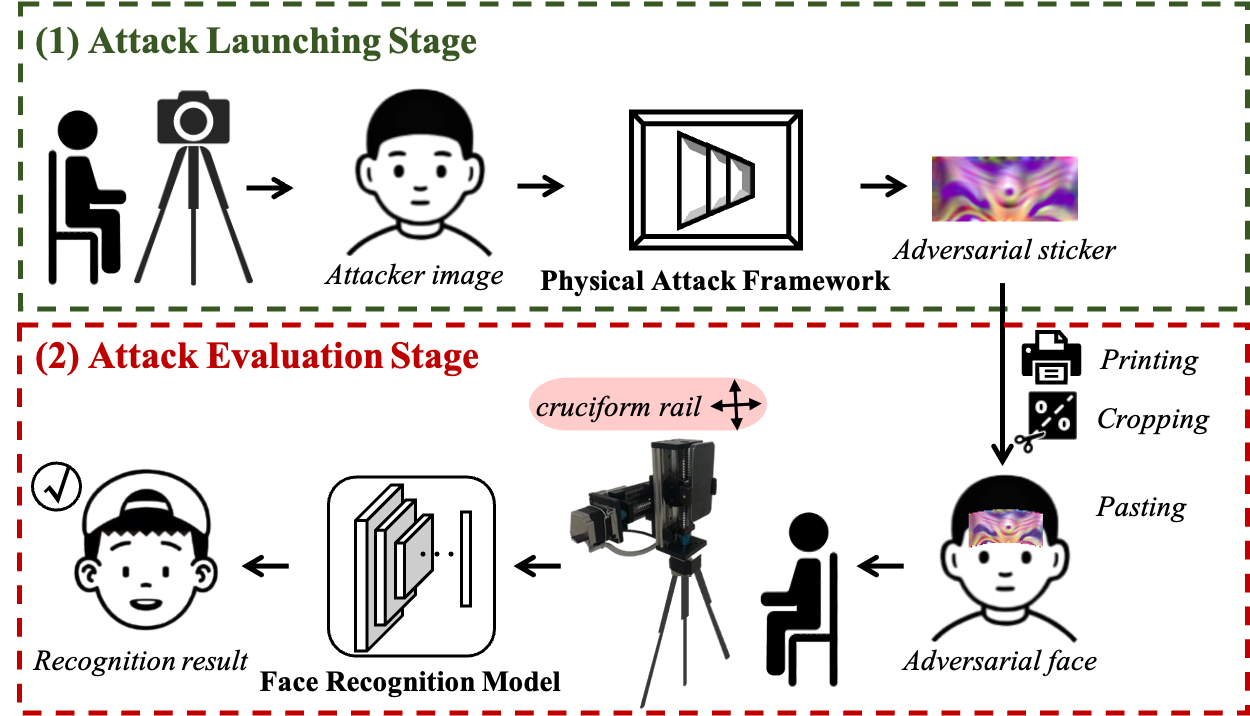} 
	\caption{The whole pipeline of the proposed testing protocol.}
	\label{Fig.protocol}
\end{figure}

The constructed testing protocol is shown in Figure~\ref{Fig.protocol}, which contains two phases: (1) {\it attack launching stage} for collecting facial images and generating adversarial stickers, and (2) {\it attack evaluation stage} for evaluating attacking performance under various physical conditions.

\begin{table}[t]
\centering
\resizebox{0.43\textwidth}{!}{
\begin{small}
\begin{tabular}{lccccccc}
\toprule
Methods & Experimenters & D   & I   & Illus & FaceVars & Poses    & Images \\ \midrule
Advhat  & 10       & $\checkmark$ & $\times$  & 3     & 1        & 8        & 128     \\ 
\bf{Ours}    & \bf{10}       & \bf{$\checkmark$} & \bf{$\checkmark$} & \bf{3}     & \bf{4}        & \bf{$\sim$35} & \bf{5880}       \\ \bottomrule
\end{tabular}
\end{small}}
\caption{Statistic comparison of physical evaluation cases between Advhat \cite{komkov2019advhat} and our PadvFace. (\scriptsize{D: dodging attacks; I: impersonation attacks; Illus/FaceVars/Poses: number of illumination/facial/pose variations; Images: number of evaluated images.})}
\label{tab:num_cases}
\vskip -0.1in
\end{table}

\textbf{Attack Launching Stage.} 
For each experimenter, we first take 4 videos under the normal light with 4 different facial variations: {\it happy, sad, neutral}, and {\it mouth-open}.
We use iPhone-12 to take 1920$\times$1080 resolution videos and each video lasts about 3 $\sim$ 5 seconds.
The camera is placed in front of the experimenter with a distance of 50cm and the experimenter is asked to sit steady without pose variations.
Then, we randomly sample a single frame for each video as $\mathcal{S}_{\bm{x}}$, which would be taken as inputs of attack models for generating robust adversarial stickers. 

\textbf{Attack Evaluation Stage.} 
The stickers generated in the launching stage are firstly printed by Canon Generic Plus PCL6 and then wore to deceive face recognition systems.
We consider complicated environmental variations for the evaluations of this stage, including
(1) {\it different head poses}.
Existing methods usually acquire this by asking the experimenter to make certain pose changes, which is hard to control and cannot lead to a fair comparison between different methods due to the poor repeatability.
To alleviate this issue, we customize a cruciform rail to make accurate movements for reducing the potential effects by uncontrolled experimenter movements.
The experimenter is asked to sit in front of the cruciform rail with a distance of 50cm, and the cruciform rail carries the camera and moves in four directions (up, down, left, and right) sequentially to capture facial images with different poses. 
As a result, we can obtain the facial images with fine-controlled pose variations. 
(2) {\it different lighting conditions.} To imitate the illumination variations, we select a room without any extra window and use a KN-18C annular light as the light source, which is also placed in front of the experimenter with a distance of 50cm. 
We select three different light intensities, refer to dark/normal/light respectively, and examples of face images under different lighting conditions are provided in \textit{Appendix}.
(3) {\it facial variations.} As considered in the attack launching stage, we also involve four facial variations: happy/sad/neutral/mouth-open.

For each experimenter with a certain sticker, we take six videos: (a) three videos under the dark/normal/light lighting conditions with the neutral expression, and (b) three videos with the happy/sad/mouth-open variations under the normal illumination. 
The movement ranges of the cruciform rail are kept same for each video, \ie, left-right angles $(-{{5}^\circ},+{{5}^ \circ})$ and the up-down angles $(-{{2}^\circ},+{{2}^ \circ})$.
Afterwards, about 35 frames are captured from each video as the adversarial images $\bm{x}^{adv}$.
For a better evaluation, we also collect six videos for each experimenter without any stickers, \ie, benign faces ${{\bm{x}}^{b}}$.

We take the `neutral' facial image captured from the attack launching round as an anchor ${\bm{x}^a}$, and calculate the benign cosine similarity ${co{s_{b}}}=cos({{\bm{x}}^b},{\bm{x}^a})$ and the adversarial cosine similarity ${co{s_{adv}}}=cos({{\bm{x}}^{adv}},{\bm{x}^a})$.
Instead of the attack success rate, taking the cosine similarity as the metric would lead to more precise evaluation without the effects caused by the threshold setting.

The statistics of the physical evaluation cases are presented in Table~\ref{tab:num_cases}, where we also tabulate those of Advhat \cite{komkov2019advhat} for comparison.
It is worth noting that the evaluation of physical attacks is much more challenging than that of digital attacks, making the number of experimenters or attack cases relatively small.
Yet our evaluations are already on the largest scale compared to existing works.

\textbf{Experimental Setting.}
We take the state-of-the-art ArcFace model\footnote{https://github.com/deepinsight/insightface/wiki/Model-Zoo} trained on the large-scale dataset MS1MV2 as our attacked face recognition model, which has 99.77\% benign recognition accuracy on the LFW benchmark \cite{huang2008labeled}.
The default sticker size is 400$\times$900 in pixels, obtained 13cm$\times$5.8cm in the physical world. 
All experiments are conducted with the TensorFlow platform and a NVIDIA Tesla P40 GPU.
The detailed setting of the proposed CAA algorithm is provided in \textit{Appendix}.

In the following, we denote the proposed method that considers all physical variations with the CAA optimizer as \textbf{PadvFace-$F$} and introduce its two variants: 
\textbf{PadvFace-$B$} that does not involve facial variations in $\mathcal{S}_{\bm{x}}$ and illumination transformations in $\mathcal{T}^{B}$, and \textbf{PadvFace-$S$} realized by further substituting the standard EOT optimizer for the CAA optimizer in \textbf{PadvFace-$B$}.

\subsection{Evaluations of Physical Attacks}

\begin{figure}[t] 
	\centering
	\includegraphics[width=0.38\textwidth]{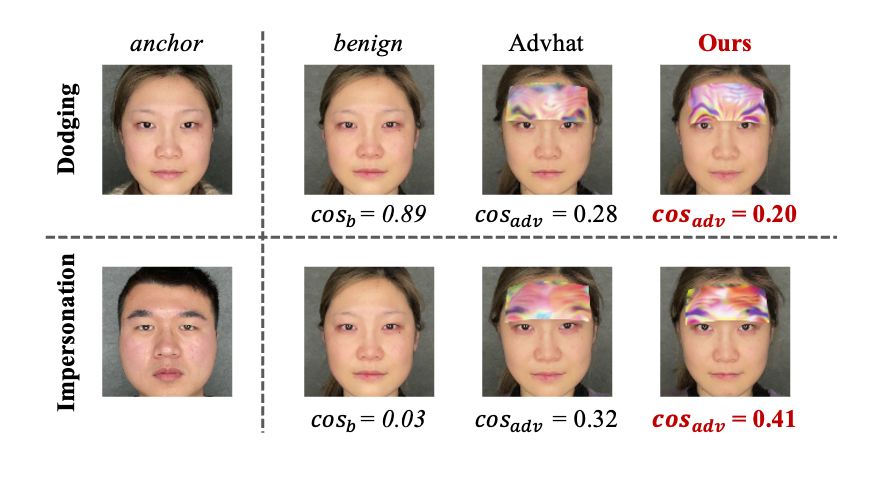} 
	\caption{Face examples of physical dodging and impersonation attacks on Advhat \vs\ Ours (PadvFace-\textit{B}).}
	\label{Fig.Comparison_phy}
\end{figure}

In this section, we evaluate the proposed PadvFace with dodging and impersonation attacks in the physical world. 
And we take Advhat \cite{komkov2019advhat} as the main baseline, which is the most comparative sticker-based attack method for our evaluations on large-scale face recognition.
Since Advhat did not capture the internal facial variations and illumination variations during the optimization, we align this setting and adopt the {PadvFace-$B$} for a fair comparison. 
Nevertheless, our {PadvFace-$B$} still keeps the D2P module and CAA optimization algorithm, which are the two main differences in contrast to Advhat. 
In consequence, we evaluate both methods with the `neutral' expression under the `normal' light and Fig.~\ref{Fig.Comparison_phy} provides some attack examples.
The numerical results of dodging and impersonation attacks on 10 experimenters are reported in Table~\ref{tab:compare_sota}, where $cos_b$ and $cos_{adv}$ denote the benign and the adversarial cosine similarity, respectively. For dodging attacks, lower $cos_{adv}$ denotes better performance, while for impersonation attacks, higher $cos_{adv}$ denotes better performance.

\begin{table}[t]
\centering
\resizebox{0.48\textwidth}{!}{
\begin{tabular}{ccccccccc}
\toprule
\multicolumn{4}{c}{\bf{Dodging Attack ($\downarrow$)}}    &  & \multicolumn{4}{c}{\bf{Impersonation Attack ($\uparrow$)}} \\ \cline{1-4}\cline{6-9}\specialrule{0em}{1.5pt}{1.5pt}
\multicolumn{1}{c}{ID} & \multicolumn{1}{c}{$\underset{\left(cos_{b}\right)}{\rm benign}$} & \multicolumn{1}{c}{$\underset{(\cos_{adv})}{\rm Advhat}$} & \multicolumn{1}{c}{$\underset{(\cos_{adv})}{\rm \textbf{Ours}}$} &  & \multicolumn{1}{c}{ID} & \multicolumn{1}{c}{$\underset{(\cos_{b})}{\rm benign}$} & \multicolumn{1}{c}{$\underset{(\cos_{adv})}{\rm Advhat}$} & \multicolumn{1}{c}{$\underset{(\cos_{adv})}{\rm \textbf{Ours}}$} \\ \midrule
01                       & 0.91                          & 0.29                        &\bf{0.27}                      &  & 01$\rightarrow$02                & 0.16                          & 0.28                        & \bf{0.46}                       \\ 
02                       & 0.94                          & 0.43                        &\bf{0.33}                      &  & 02$\rightarrow$03                & 0.09                          & 0.26                        & \bf{0.35}                       \\
03                       & 0.91                          & \bf{0.33}                   &\bf{0.33}                 &  & 03$\rightarrow$04                & {-0.08}                         & 0.17                        & \bf{0.21}                       \\
04                       & 0.88                          & 0.42                        &\bf{0.25}                      &  & 04$\rightarrow$09                & {0.12}                          & 0.19                        & \bf{0.21}                       \\
05                       & 0.94                          & 0.60                        &\bf{0.51}                      &  & 05$\rightarrow$04                & {0.04}                          & 0.21                        & \bf{0.24}                       \\ 
06                       & 0.93                          & 0.38                        &\bf{0.32}                      &  & 06$\rightarrow$07                & {0.10}                          & 0.26                        & \bf{0.29}                       \\
07                       & 0.93                          & 0.34                        &\bf{0.31}                      &  & 07$\rightarrow$08                & {0.20}                          & 0.33                        & \bf{0.37}                       \\ 
08                       & 0.95                          & 0.32                        &\bf{0.26}                      &  & 08$\rightarrow$09                & {0.12}                          & 0.24                        & \bf{0.33}                       \\ 
09                       & 0.93                          & 0.37                        &\bf{0.28}                      &  & 09$\rightarrow$01                & {0.09}                          & 0.19                        & \bf{0.19}                       \\ 
10                       & 0.89                          & 0.28                        &\bf{0.20}                      &  & 10$\rightarrow$03                & {0.03}                          & 0.32                        & \bf{0.41}                      \\\hline
Average & 0.92 & 0.38 & \bf{0.31} & & Average & 0.09 & 0.24 &\bf{0.30}\\
\bottomrule
\end{tabular}}
\caption{Results of Advhat and Ours (PadvFace-\textit{B}) on dodging and impersonation attacks. 
Best results are in bold. 
Each ID pair in impersonation attacks indicates `attacker$\rightarrow$victim'.}
\label{tab:compare_sota}
\end{table}

The results of dodging attacks are shown in the left of Table~\ref{tab:compare_sota}. Compared to the benign similarity without any sticker, our proposed method achieves significantly lower adversarial cosine similarities, demonstrating its superior dodging attack performance. 
For example, for ID=01, after wearing the adversarial sticker, the cosine similarity drops from 0.91 to 0.27,
posing a serious threat to the physical-world face recognition system. 
Furthermore, compared with Advhat, we also obtain significant performance improvements under most cases. 
On the average performance of all 10 experimenters, our method achieves 0.31 adversarial cosine similarity while that of Advhat is only 0.38, leading to 18.4\% relative improvement. 

As for impersonation attacks, we randomly select a single victim anchor from 10 experimenters for each attacker, obtaining 10 `attacker$\rightarrow$victim' pairs. 
The evaluations are shown in the right of Table~\ref{tab:compare_sota}.
As shown in the Table, our proposed method can significantly increase the cosine similarities of two different identities by wearing the adversarial stickers under all cases. 
For instance, for the attacking pair (10 $\rightarrow 03$), the adversarial cosine similarity increases from 0.03 to 0.41. 
Furthermore, we also obtain better attack performance than Advhat on most attacking pairs. 
On the average of all 10 cases, our method leads to an average cosine similarity of 0.30 while that of Advhat is 0.24, the relative performance improvement over Advhat is 25\%.
In addition, compared with dodging attacks, impersonation attacks that aim to deceive a specific identity are generally more difficult.

We also conduct the paired t-test further and observe that our PadvFace-B is significantly better than Advhat with p=0.005 for both dodging and impersonation attacks.
In summary, the superior performance of dodging and impersonation attacks on PadvFace-$B$ over Advhat demonstrates the effectiveness of the developed D2P module and CAA optimizer. 

\begin{table}[b]
\centering
\resizebox{0.40\textwidth}{!}{
\begin{small}
\begin{tabular}{cccc}
\toprule
\multicolumn{4}{c}{\bf{Dodging Attack($\downarrow$)}}                                                           \\ \hline\specialrule{0em}{1.5pt}{2pt}
ID                     & 04                    & 07                    & 09                    \\ \hline\specialrule{0em}{1pt}{1pt}
Dark                   & 0.81 $|$ 0.48 $|$ \bf{0.38}    & 0.92 $|$ 0.38 $|$ \bf{0.34}    & 0.91 $|$ 0.56 $|$ \bf{0.48}    \\ 
Normal                 & 0.83 $|$ 0.50 $|$ \bf{0.38}    & 0.95 $|$ 0.44 $|$ \bf{0.38}    & 0.93 $|$ 0.50 $|$ \bf{0.46}    \\ 
Light                  & 0.81 $|$ 0.49 $|$ \bf{{0.36}}    & 0.96 $|$ 0.39 $|$ \bf{{0.29}}    & 0.90 $|$ 0.49 $|$ \bf{{0.43}}    \\\hline\specialrule{0em}{0.5pt}{0.5pt} 
Average           & 0.82 / 0.49 / \textbf{0.37}    & 0.94 / 0.40 / \textbf{0.34}    & 0.92 / 0.52 / \textbf{0.45}    \\\specialrule{0em}{1.5pt}{1.5pt} 
\toprule
\multicolumn{4}{c}{\bf{Impersonation Attack($\uparrow$)}}                            \\ \hline\specialrule{0em}{1.5pt}{2pt}
ID                     & 07$\rightarrow$08                 & 08$\rightarrow$09                 & 09$\rightarrow$10                 \\ \hline\specialrule{0em}{1pt}{1pt}
Dark                   & 0.21 $|$ 0.36 $|$ \bf{0.37}    & 0.10 $|$ 0.33 $|$ \bf{0.39}    & 0.03 $|$ 0.32 $|$ \bf{0.40}    \\ 
Normal                 & 0.20 $|$ 0.37 $|$ \bf{{0.40}}    & 0.12 $|$ 0.33 $|$ \bf{0.37}    & 0.03 $|$ 0.33 $|$ \bf{{0.41}}    \\ 
Light                  & 0.22 $|$ 0.36 $|$ \bf{0.39}    & 0.14 $|$ 0.37 $|$ \bf{{0.42}}    & 0.04 $|$ 0.37 $|$ \bf{{0.41}}    \\\hline \specialrule{0em}{0.5pt}{0.5pt}
Average                   & 0.21 / 0.36 / \textbf{0.39}    & 0.12 / 0.35 / \textbf{0.40}    & 0.03 / 0.34 / \textbf{0.41}     \\ \bottomrule
\end{tabular}
\end{small}}
\caption{Comparison with the neutral expression under illumination variations. Metric is `benign ($cos_{b}$) $|$ PadvFace-\textit{B} ($cos_{adv}$) $|$ PadvFace-\textit{F} ($cos_{adv}$)'. Best results are in bold.}
\label{tab:illus}
\end{table}

\subsection{Experiments of Environmental Variations}
With the proposed standardized testing protocol, we can make more quantitative analyses of environmental variations. 
In this section, we analyze in-depth the impact of internal facial variations and illumination variations. 
We compare the performance of PadvFace-\textit{B} and PadvFace-\textit{F} under both dodging and impersonation attacks. 

The evaluation results under different lighting conditions are given in Table~\ref{tab:illus}, where all experimenters are asked to keep the neutral expression during the attack evaluation stage. 
We have the following observations: (1) the attacking performance varies under different illumination conditions,
and (2) PadvFace-\textit{F} that incorporates the illumination transformations in $\mathcal{T}^B$ consistently outperforms PadvFace-\textit{B} that learned with a fixed normal illumination under all cases. Specifically, for ID=09, PadvFace-\textit{F} has 12.5\% performance improvement than PadvFace-\textit{B} (0.52 \vs\ 0.45) for dodging attacks, and 19.9\% performance improvement (0.34 \vs\ 0.41) for impersonation attacks.

The evaluations under internal facial variations are presented in Table~\ref{tab:exps}, where all adversarial images are collected under the normal illumination. 
It can be observed that when attackers make different facial variations, the attack performance would vary as well. 
This point can be further verified by the superior performance of PadvFace-\textit{F} over PadvFace-\textit{B}, where PadvFace-\textit{F} involves facial variations during the optimization but PadvFace-\textit{B} learns only based on neutral expression. Specifically, for ID=09, PadvFace-\textit{F} achieves 9.4\% dodging attack performance improvement (0.53 \vs\ 0.48) and 13.0\% impersonation attack performance improvement (0.31 \vs\ 0.35) compared with PadvFace-\textit{B}.
These experimental results demonstrate that considering internal facial variations and illumination variations during the attacking process can largely boost the robustness of learned adversarial stickers.

\begin{table}[t]
\centering
\resizebox{0.40\textwidth}{!}{
\begin{small}
\begin{tabular}{cccc}
\toprule
\multicolumn{4}{c}{\bf{Dodging Attack ($\downarrow$)}}                                      \\ \hline\specialrule{0em}{1.5pt}{3pt}
ID         & 04                 & 07                 & 09                 \\ \hline\specialrule{0em}{1pt}{1pt}
Happy      & 0.80 $|$ 0.41 $|$ \textbf{0.32} & 0.94 $|$ 0.40 $|$ \textbf{{0.34}}  & 0.90 $|$ 0.51 $|$ \textbf{0.49} \\ 
Sad        & 0.66 $|$ 0.29 $|$ \textbf{{0.18}} & 0.93 $|$ 0.43 $|$ \textbf{0.35} & 0.79 $|$ 0.47 $|$ \textbf{{0.43}} \\ 
Neutral    & 0.83 $|$ 0.50 $|$ \textbf{0.38} & 0.95 $|$ 0.44 $|$ \textbf{0.38} & 0.93 $|$ 0.50 $|$ \textbf{0.46} \\ 
Mouth-open & 0.78 $|$ 0.47 $|$ \textbf{0.32} & 0.91 $|$ 0.45 $|$ \textbf{0.35} & 0.82 $|$ 0.65 $|$ \textbf{0.54} \\ \hline \specialrule{0em}{0.5pt}{0.5pt}
Average       & 0.77 / 0.42 / \textbf{0.30}     & 0.93 / 0.43 / \textbf{0.35}     & 0.86 / 0.53 / \textbf{0.48}     \\\specialrule{0em}{1.5pt}{1.5pt}
\toprule\specialrule{0em}{1.5pt}{2pt}
\multicolumn{4}{c}{\bf{Impersonation Attack ($\uparrow$)}}                                \\ \hline\specialrule{0em}{1.5pt}{3pt}
ID         & 07$\rightarrow$08              & 08$\rightarrow$09              & 09$\rightarrow$10              \\ \hline\specialrule{0em}{1pt}{1pt}
Happy      & 0.22 $|$ 0.36 $|$ \textbf{0.39} & 0.10 $|$ 0.32 $|$ \textbf{0.39} & 0.02 $|$ 0.34 $|$ \textbf{0.37} \\
Sad        & 0.18 $|$ 0.22 $|$ \textbf{0.31} & 0.12 $|$ 0.36 $|$ \textbf{{0.40}} & 0.07 $|$ 0.32 $|$ \textbf{0.35} \\
Neutral    & 0.20 $|$ 0.37 $|$ \textbf{{0.40}} & 0.12 $|$ 0.33 $|$ \textbf{0.37} & 0.03 $|$ 0.33 $|$ \textbf{{0.41}} \\
Mouth-open & 0.21 $|$ 0.34 $|$ \textbf{0.38} & 0.15 $|$ 0.34 $|$ \textbf{0.39} & 0.06 $|$ 0.26 $|$ \textbf{0.29} \\ \hline \specialrule{0em}{0.5pt}{0.5pt}
Average       & 0.20 / 0.32 / \textbf{0.37}     & 0.12 / 0.34 / \textbf{0.39}     & 0.04 / 0.31 / \textbf{0.35}     \\ \bottomrule
\end{tabular}
\end{small}}
\caption{Comparison with internal facial variations under the normal illumination. Metric is `benign ($cos_{b}$) $|$ PadvFace-\textit{B} ($cos_{adv}$) $|$ PadvFace-\textit{F} ($cos_{adv}$)'. Best results are in Bold.}
\label{tab:exps}
\end{table}

\subsection{Ablation Study}
\begin{figure}[b] 
	\centering
	\includegraphics[width=0.25\textwidth]{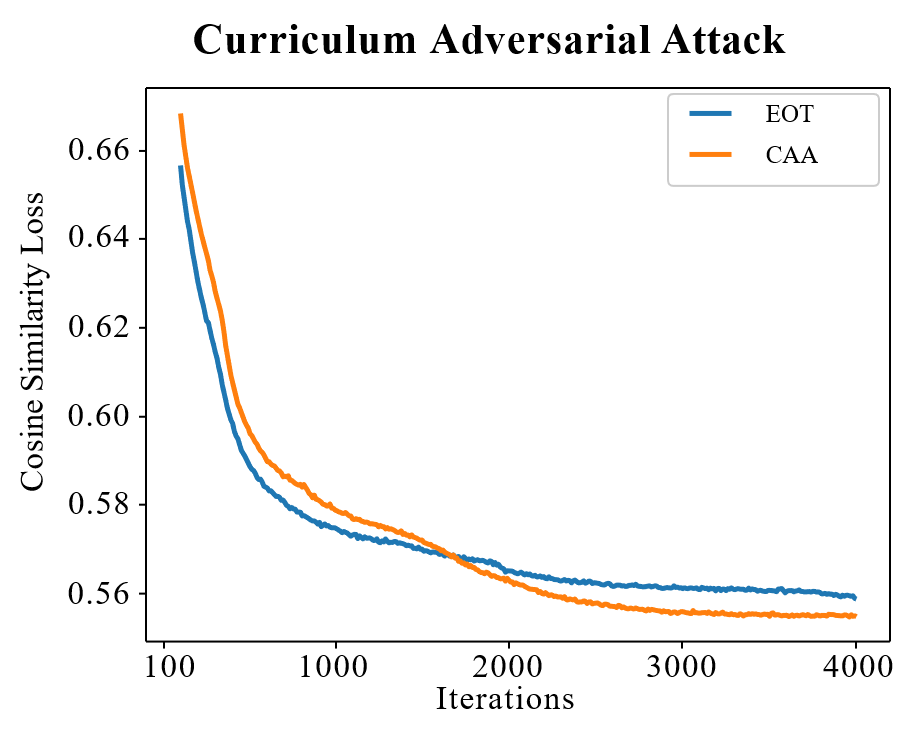} 
	\caption{Convergence of dodging attack performance with the proposed CAA \vs\ existing EOT optimization algorithms.}
	\label{Fig.CAA}
\end{figure}

\textbf{D2P Module}.
We utilize PadvFace-\textit{S} for evaluating the effectiveness of the proposed D2P module. 
We randomly select three experimenters and conduct dodging attacks in physical world.
The evaluation results are presented in Table~\ref{tab:d2p}, where `w/ D2P' refers to PadvFace-\textit{S} and `w/o D2P' refers to the variant of PadvFace-\textit{S} without the D2P module.
The benign and adversarial images are captured under the `neutral' expression and `normal' illumination condition.
As can be observed, the D2P module can significantly benefit the performance of physical attacks for all three experimenters, demonstrating the effectiveness of modeling the chromatic aberration induced by printers and cameras.
\begin{table}[h]
\centering
\resizebox{0.28\textwidth}{!}{
\begin{footnotesize}
\begin{tabular}{p{1cm}<{\centering}p{1.25cm}<{\centering}p{1.25cm}<{\centering}p{1.25cm}<{\centering}}
\toprule
ID & benign & w/o D2P & w/ D2P \\ \midrule
01 & 0.87     & 0.29    & \bf{0.22}   \\ 
02 & 0.95     & 0.38    & \bf{0.32}   \\ 
10 & 0.89     & 0.19    & \bf{0.16}   \\ \bottomrule
\end{tabular}
\end{footnotesize}
}
\caption{Ablation study of D2P module.}
\label{tab:d2p}
\end{table}

\smallskip
\noindent
\textbf{CAA Algorithm.}
In Fig.~\ref{Fig.CAA}, we plot the tendency curves of the cosine similarity loss with a certain experimenter, to explore the difference in optimizing process between CAA and EOT. CAA refers to PadvFace-\textit{F} and EOT refers to the variant of PadvFace-\textit{F} by replacing the CAA optimizer with the EOT optimizer. The cosine similarity loss at each iteration in Fig.~\ref{Fig.CAA} is calculated by the average of the adversarial cosine losses of 400 randomly sampled transformations of $\mathcal{K}$. As can be expected, learning with easy physical-world conditions causes a relatively slower convergence rate in the early optimization stage compared with the EOT optimizer. 
However, as the iteration increases, more and more complex physical-world conditions are involved and the proposed CAA algorithm leads to better performance with lower cosine similarity loss at the end of the learning process.

\begin{table}[t]
\centering
\resizebox{0.36\textwidth}{!}{
\begin{small}
\begin{tabular}{lccc}
\toprule
Target Models           & benign & Advhat & \bf{Ours} \\ \midrule
CosFace \cite{yang2020delving}     & 0.30   & 0.34   & \bf{0.38} \\ 
MobileFace \cite{yang2020delving}      & 0.17   & 0.26   & \bf{0.29} \\ 
CurriculumFace \cite{cai2018curriculum} & 0.06   & 0.14   & \bf{0.16} \\ \bottomrule
\end{tabular}
\end{small}}
\caption{Transferability of impersonation attacks.}
\label{tab:transfer}
\end{table}
\subsection{Discussions}
{\textbf{Inconspicuousness.} For physical attacks, the {inconspicuousness} aims to make adversarial perturbations unnoticed. However, based on our experiments, attacking physical-world face recognition is intrinsically hard, and the current attack performance is far from satisfactory even without the inconspicuousness constraint. Thus, in this paper, we primarily focus on efficiently modeling the complicated physical-world conditions in attacking face recognition, and leaving the inconspicuousness of the adversarial stickers to future work.
}

\smallskip
\noindent
\textbf{Transferability.} We evaluate the attack robustness of the proposed PadvFace when transferring to other face recognition models. The average cosine similarity of 10 experimenters are provided in Table~\ref{tab:transfer}, with ArcFace as the source model. As can be observed, our proposed method can achieve consistently better performance than Advhat on all target models. Note that we do not specifically impose constraints on the transferability of PadvFace. Nevertheless, we believe that it could be possible to introduce the advanced study from another branch of adversarial attacks, i.e., transfer attacks, to further improve the transferability.
\section{Conclusion}
In this work, we study the adversarial vulnerability of physical-world face recognition by sticker-based adversarial attacks. For robust adversarial attacks, we analyze in detail the complicated physical-world condition variations in attacking face recognition and propose a novel physical attack method that considers and models these variations. We further propose an efficient curriculum adversarial attack algorithm that gradually learns the sticker from easy to complex physical-world variations. We construct a standardized testing protocol for facilitating the fair evaluation of physical attacks on face recognition. Extensive experimental results demonstrate the effectiveness of the proposed method for dodging and impersonation physical attacks.

\bibliography{aaai22}

\maketitle
\newpage
\appto\appendix{\setcounter{secnumdepth}{0}}
\appendix
\section{Appendix}
\hspace*{\fill}
\renewcommand\thefigure{\Alph{section}\arabic{figure}} 
\renewcommand\thetable{\Alph{section}\arabic{table}}
\subsection{A.1~~Details of Digital-to-Physical (D2P) Module}
\label{sec:sup-d2p}

As analyzed in Sec.~3.3, the D2P module aims to imitate the chromatic aberration induced by printers and cameras during physical attacks. It is realized by a two-layer MLP with 100 hidden nodes to learn a 1:1 color mapping from the digital space to the physical space. In the following, we first report the details of producing color palettes and then provide the training details of the MLP. Furthermore, we provide the evaluation results of the D2P module on adversarial stickers.
\begin{figure}[b] 
	\centering
	\includegraphics[width=0.5\textwidth]{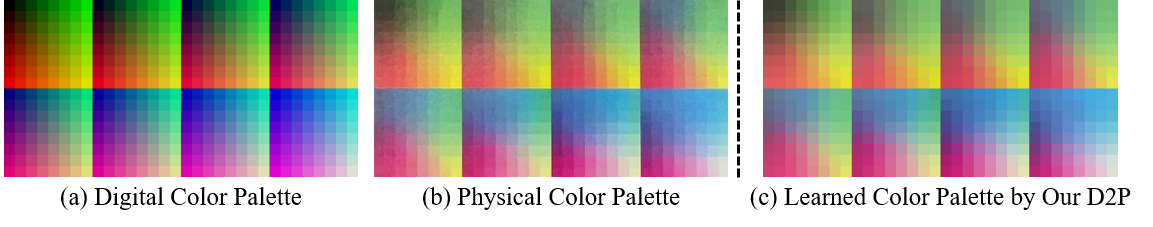}
	\caption{Color palettes.}
	\label{Fig.colors}
\end{figure}

\textbf{Color Palettes.} 
For the learning of the MLP, we need to capture both RGB colors in the digital space and their corresponding values after printing and photographing in the physical world.
To this end, we first construct a set of color anchors $A^d$ in the digital space. 
Since we can not enumerate all colors in RGB space, we only make a subset of it.
Specifically, our color anchors $A^d$ consist of 512 colors generated by sampling Red, Green, and Blue colors, respectively. 

To better capture the corresponding colors after printing and photographing $A^d$ in the physical world, we reshape $A^d$ to the size of 16$\times$32 and replicate each color anchor of $A^d$ to 40$\times$40 pixel square, leading to the digital color palette with 640$\times$1280 in pixels in Fig.~\ref{Fig.colors} (a). 
Then, we print the digital color palette and photograph it under the normal illumination with the distance of 50cm in our testing environment (refers to Sec.~4.1), obtaining the physical color palette in Fig.~\ref{Fig.colors} (b). In addition, we average the pixel square of $\frac{h}{{16}}\times\frac{w}{{32}}$, where $h$ and $w$ denote the height and the width of (b), resulting in $A^p$ with the size of 16$\times$32 as the corresponding colors of $A^d$ in the physical world. 

\textbf{MLP Training Details.} Taking the digital color anchors $A^d$ as the input and its physical counterpart $A^p$ as the ground truth, we train our MLP through Adam optimizer with 100,000 epochs. The initial learning rate is 0.01 that decays by the factor of 10 at the epochs of 50,000 and 70,000, respectively. As a result, we can obtain the learned color palette by D2P in Fig~\ref{Fig.colors} (c), which only has the Mean Square Error (MSE) of 0.0001 with the ground truth $A^p$ of (b).

\begin{figure}[t] 
	\centering
	\includegraphics[width=0.5\textwidth]{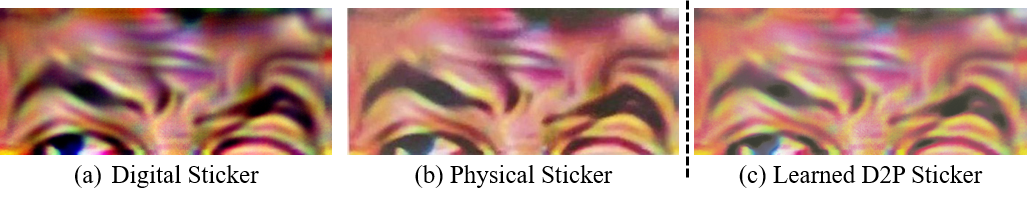}
	\caption{Examples of adversarial stickers.}
	\label{Fig.stickers}
\end{figure}

\textbf{Further Evaluation.} This part further evaluates the performance of the learned D2P module on adversarial stickers. As shown in Fig.~\ref{Fig.stickers}, given an arbitrary digital adversarial sticker in (a), we first print and photograph it in our testing environment, obtaining the physical sticker in (b). In the meantime, the digital sticker is also fed into the D2P module to obtain the learned D2P sticker in (c). We test Peak Signal-to-Noise Ratio (PSNR), Mean Structural Similarity (MSSIM), and MSE, between pairs `Digital-to-Physical' stickers, \ie, (a) and (b),  as well as `D2P-to-Physical' stickers, \ie, (c) and (b), in Fig.~\ref{Fig.stickers} for evaluating the performance of trained MLP. Corresponding results are reported in Table~\ref{tab:metric-sticker}. Note that higher MSSIM and lower PSNR and MSE denote better results. As can be observed, the adversarial sticker learned by our D2P is closer to the physical adversarial sticker than the digital one. Therefore, the proposed D2P module can effectively address the chromatic aberration induced by printers and cameras during physical attacks.
\begin{table}[h]
\centering
\begin{small}
\begin{tabular}{lccc}
\toprule
Sticker Metrics     & PSNR (dB)$\downarrow$  & MSSIM$\uparrow$ & MSE$\downarrow$   \\ \midrule
Digital-to-Physical & 18.27 & 0.55  & 0.014 \\ 
D2P-to-Physical     & \bf{21.58} & \bf{0.62}  & \bf{0.006} \\ \bottomrule
\end{tabular}
\end{small}
\caption{Performance of D2P module over adversarial sticker pairs. Higher MSSIM $\uparrow$ and lower PSNR$\downarrow$ and MSE$\downarrow$ denote better results. Best results are shown in bold.}
\label{tab:metric-sticker}
\end{table}

\begin{table}[b]
\centering
\begin{small}
\resizebox{0.35\textwidth}{!}{%
\begin{tabular}{llcc}
\toprule
Module & Variations                                & Min     & Max    \\ \midrule
\multirow{4}{*}{$\mathcal{T}^A$}     & Parabolic angle         & -$2^\circ$      & +$2^\circ$     \\ 
       & Parabolic rate & -2$\times$$10^{-4}$  & +2$\times$$10^{-4}$ \\ 
       & Rotation                                  & -$1^\circ$      & +$1^\circ$     \\
       & Translation                               & -1      & +1     \\ \hline\specialrule{0em}{0.5pt}{0.5pt} 
\multirow{5}{*}{$\mathcal{T}^B$}     & Rotation                                  & -$3^\circ$      & +$3^\circ$     \\ 
       & Scaling                                   & 0.94 & 1.06  \\ 
       & Translation                               & -2      & +2     \\
       & Contrast                                  & 0.5     & 1.1    \\
       & Brightness                                & 0.05    & 0.1    \\\bottomrule
\end{tabular}}
\end{small}
\caption{Transformation parameters, where each parameter is randomly sampled at equal intervals from the specified range. `Parabolic angle' and `Parabolic rate' are integrated as `Parabolic transformations'.}
\label{tab:trans}
\end{table}

\subsection{A.2~~Details of Curriculum Adversarial Attack (CAA) Algorithm}
\label{sec:sup-caa}
For the proposed curriculum adversarial attack method, the weight of TV loss is set to $\alpha=10^{-5}$ in model (4). 
More specifically, in Algorithm 1, for the updating of $\bm{\delta}$, the learning rate is set to 0.02 with the momentum of 0.95. 
We adopt three curriculum learning stages, \ie, $T=3$ in Algorithm 1.
The number of inner iteration $E$ is set to 2000/2000/3000 for each curriculum stage, respectively.
Moreover, at the beginning of each curriculum learning stage $t$, we determine the curriculum parameter $\lambda_t$ as follows:
let ${\beta _t} = \frac{1}{n}\sum\nolimits_{i = 1}^n {{p_i}}$ denote the curriculum proportion, where $p_i = 1- \mathcal{L}_{sim, k_i}/\lambda_t$.
We set the curriculum proportion ${\beta_1},{\beta_2},{\beta_3}$ to $0.5$, 0.8 and 1.0, respectively. 
Besides, each inner updating (step3 and step4 of Algorithm 1) is performed based on a randomly sampled mini-batch of size 32 from $\mathcal{K}$ for the training efficiency.

In addition, we provide the settings of the transformation parameters of $(\mathcal{S}_{\bm{x}}, \mathcal{T}^A, \mathcal{T}^B, \bm{v})$. 
Note that $\mathcal{S}_{\bm{x}}$ contains four facial variations (\ie, happy/neutral/sad/mouth-open) and $\bm{v}\sim \mathcal{N}(\mu,\sigma)$ is a random Gaussian noise with $\mu=0$ and $\sigma=0.02$.
The transformation parameters of $\mathcal{T}^A$ and $\mathcal{T}^B$ are given in Table~\ref{tab:trans}.

\subsection{A.3~~Examples of Three Light Conditions}
\begin{figure}[h] 
	\centering
	\includegraphics[width=0.4\textwidth]{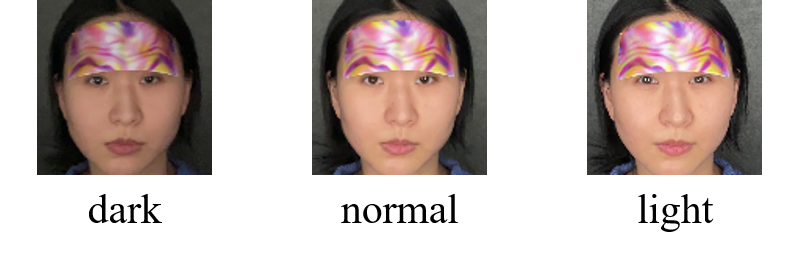}
	\caption{Examples of adversarial faces under three real-world light conditions.}
	\label{Fig.illus_samples}
\end{figure}

We provide examples of adversarial faces under three real-world light conditions (\ie, dark/normal/light) used in our experiments in Fig.~\ref{Fig.illus_samples}. All facial images come from our collected dataset following the proposed testing protocol.

\end{document}